\documentclass[twocolumn]{IEEEtran}

\usepackage{graphicx}
\usepackage{amsmath}
\usepackage[margin=1in]{geometry}
\usepackage{multirow}
\usepackage{graphicx}
\usepackage{amsmath}
\usepackage{xcolor}
\usepackage{colortbl}
\usepackage{graphicx} % For including figures
\usepackage{lipsum} % For generating dummy text
\usepackage{multicol} % For multiple columns
\usepackage{placeins}
\usepackage{graphicx}
\usepackage{float}
\usepackage[misc]{ifsym}
\usepackage{amsmath}
\usepackage[T1]{fontenc}
\usepackage{aecompl}
\usepackage{diagbox}
\usepackage{authblk}
\usepackage{bbding}
\begin{document}

\newcommand{\customDataRow}[9]{%
  #1 & #2 & #3 & #4 & #5 & #6 & #7 & #8 & #9  \\
}
% \title{Uncertainty Estimation for Multi-modal Fusion Object Detection in Autonomous Driving}

% \author{Your Name}

% \maketitle

\title{MMLF: Multi-modal Multi-class Late Fusion for Object Detection with Uncertainty Estimation

\thanks{ 
\Envelope Yang Zhao is the corresponding author.

*Qihang Yang is with Glasgow College, University of Electronic Science and Technology of China, Chengdu, China, 611731(email: qihang.yang@std.uestc.edu.cn)

*Yang Zhao is with the School of Automation Engineering, University of Electronic Science and Technology of China, Chengdu, China, 611731(email: yzhao@uestc.edu.cn)

*Hong Cheng is with the School of Automation Engineering, University of Electronic Science and Technology of China, Chengdu, China, 611731(email:hcheng@uestc.edu.cn)
}

}
% \author{Qihang Yang\thanks{{\Letter} Corresponding author}  \And  Yang Zhao {\Letter} \thanks{Yang Zhao: yzhao@uestc.edu.cn}}

 % \author{Qihang Yang
 % \thanks{Qihang Yang is with Glasgow College,  University of Electronic Science and Technology of China, Chengdu, China, 611731(email: qihang.yang@std.uestc.edu.cn)}
 % \thanks{{\Letter}Yang Zhao is with the School of Automation Engineering, University of Electronic Science and Technology of China, Chengdu, China, 611731(corresponding author, email: yzhao@uestc.edu.cn)
 % }
% \author{
%   Qihang Yang, \thanks{Qihang Yang is with Glasgow College,  University of Electronic Science and Technology of China, Chengdu, China, 611731(email: qihang.yang@std.uestc.edu.cn)}
%   \and  Yang Zhao}
\author{{Qihang Yang*, Yang Zhao*\Envelope, Hong Cheng*}}

\maketitle

% \title{More than one Author with different Affiliations}
% \author[a]{Author A}
% \author[a]{Author B}
% \author[a]{Author C \thanks{Corresponding author: email@mail.com}}
% \author[b]{Author D}
% \author[b]{Author E}
% \affil[a]{Department of Computer Science, \LaTeX\ University}
% \affil[b]{Department of Mechanical Engineering, \LaTeX\ University}

% % 使用 \thanks 定义通讯作者

% \renewcommand*{\Affilfont}{\small\it} % 修改机构名称的字体与大小
% \renewcommand\Authands{ and } % 去掉 and 前的逗号
% \date{} % 去掉日期

%   \maketitle

\begin{abstract}
Autonomous driving necessitates advanced object detection techniques that integrate information from multiple modalities to overcome the limitations associated with single-modal approaches. The challenges of aligning diverse data in early fusion, and the complexities along with overfitting issues introduced by deep fusion underscore the efficacy of late fusion at the decision level.  Late fusion ensures seamless integration without altering the original detector's network structure. This paper introduces a pioneering Multi-modal Multi-class Late Fusion (MMLF) method, which is designed for late fusion to enable multi-class detection. Fusion experiments conducted on the KITTI validation and official test datasets illustrate substantial performance improvements, presenting our model as a versatile solution for multi-modal object detection in autonomous driving. Moreover, our approach incorporates uncertainty analysis into the classification fusion process, which renders our model more transparent and trustworthy, providing more reliable insights into category predictions.
\end{abstract}

\section{Introduction}
In recent years, there has been a substantial increase in demand for robust and accurate object detection methods across various applications such as autonomous driving vehicles, surveillance systems, and robotics\cite{masmoudi2019object}\cite{jha2021real}\cite{xu2022object}. Nevertheless, conventional single-modal approaches rely solely on data from individual sensors like LiDAR or cameras, often failing to provide the comprehensive information necessary for accurate and reliable object detection, especially in complex and dynamic scenarios \cite{geiger2012we}. Given these limitations, there is a growing inclination in the research community towards adopting multi-modal approaches. This strategic shift aims to leverage the complementary advantages of different sensor modalities to address the drawbacks of individual sensors and enhance overall detection capabilities, offering a more robust and general solution for object detection. Multi-modal fusion provides a promising way of combining inputs to enhance the information available for object detection from multiple sensors. However, existing fusion techniques face challenges, such as the need for precise data alignment in early fusion methods increased computational demands, and limited scalability in deep fusion methods \cite{huang2202multi}.

Beyond the challenges posed by multi-modal data fusion, uncertainty analysis methods are also critically important in predictions \cite{kraus2019uncertainty}. The dynamics and complexity of real-world scenarios necessitate a deeper understanding of uncertainty related to object detection predictions. In recent years, there has been a growing recognition that depending solely on a single metric (e.g., confidence score) to evaluate the reliability of object detection is insufficient \cite{kraus2019uncertainty}\cite{he2019bounding}. This necessitates additional metrics (e.g., uncertainty) to offer a more comprehensive evaluation, especially in critical applications such as ensuring the safety of autonomous driving.
By integrating uncertainty analysis into the fusion process, our approach not only enhances the accuracy of the fusion process but also provides valuable insights into the uncertainty associated with the final predictions.

 To address these challenges, we propose Multi-modal Multi-class Late Fusion (MMLF) in this paper. The proposed architecture offers the following contributions:
\begin{itemize}
  \item  \textbf{Flexible Fusion of 2D and 3D Detectors:} The ability to seamlessly integrate a variety of 2D and 3D detectors, as long as they produce detections for the same set of categories.
  \item  \textbf{Late fusion of Multi-class Features:} 

The optimization of the Trusted Multi-View Classification (TMC) method \cite{TMC} for fusing pre-matched 3D and 2D candidate pairs with a non-zero Intersection over Union (IOU) provides vital information for the ultimate fusion process, completing the late-stage fusion of multi-class features.
  
%Further optimizing the fusion process involves using Trusted Multi-View Classification (TMC) techniques \cite{TMC} to hypothetically fuse pre-matched 3D and 2D candidate pairs with non-zero Intersection over Union (IOU). This approach provides significant information for the final fusion process, completing the late fusion of multi-class features.

  \item   \textbf{Uncertainty-Aware Class Fusion Analysis:} 
Quantifying the uncertainty of classification results and effectively reducing it through the fusion process.

%Leveraging the theoretical foundation of the TMC \cite{TMC}, employing mathematical methods to quantify reasonable uncertainty values for each category, and effectively reducing uncertainty through the fusion process.
\end{itemize}

The structure of this paper is as follows: Section II is a review of related research works; Section III details the proposed network structure; Section IV introduces the experimental setup, results, and analysis; Section V is a conclusion and the prospect of future work.
\section{Related Work}

\subsection{Multi-modal Fusion}
Single-modal approaches, like relying solely on visual or LiDAR data, face limitations from environmental factors or occlusions in object detection. Multi-modal fusion combines image, LiDAR, text, or audio data to enhance accuracy, robustness, and adaptability. This approach improves performance in diverse environments and can be categorized into early, deep, and late fusion methods\cite{huang2202multi}.
\subsubsection{Early fusion}
Early fusion combines information from different modalities at the raw data level. It directly merges input data from multiple sources, such as visual and LiDAR data, before processing. This approach aims to create a unified representation of the input data to feed into the object detection model. S. Vora et al. \cite{vora2020pointpainting} proposed PointPainting which operates by mapping LiDAR points onto the output of a semantic segmentation network trained solely on images, and then augmenting each point with its corresponding class scores. Li-Hua Wen et al.\cite{wen2021fast} introduce a new point feature fusion module, extracting features from raw RGB images and fusing them with their corresponding point cloud without using a backbone network. However, early fusion can be sensitive to differences in the characteristics and scales of the modalities, which may require careful preprocessing and normalization to ensure effective fusion without loss of information. Early fusion also usually involves non-generic model frameworks that may lack adaptability across diverse datasets and domains, potentially leading to suboptimal performance or requiring extensive fine-tuning for each new scenario.
\subsubsection{Deep fusion}
Deep fusion operates at a higher level by integrating features extracted from each modality within the neural network architecture. Instead of merging raw data directly, deep fusion networks learn to combine abstract representations of different modalities at various layers of the network. This approach can capture complex interactions between modalities and potentially improve detection accuracy by leveraging complementary information. One typical work is MV3D\cite{chen2017multi} which encodes the sparse 3D point cloud using a concise multi-view representation. However, deep fusion often introduces more parameters and computational complexity, leading to challenges such as overfitting and increased training time\cite{huang2020epnet}.
\subsubsection{Late fusion}
Late fusion, also known as decision-level fusion, consolidates predictions from independently trained models for each modality and integrates them during the decision-making phase. Each modality is processed separately, and its outputs are combined using fusion techniques such as averaging, voting, or more sophisticated methods like attention mechanisms. Late fusion offers greater flexibility and adaptability to new modalities since each modality model can be trained and updated independently. It also reduces the burden of dealing with modality differences during training. Figuerêdo et al. \cite{figueredo2022early} present an early detection method for depression in social media by leveraging a convolutional neural network along with context-independent word embeddings and combining Early and Late Fusion approaches. Similarly, Pang S et al. \cite{pang2020clocs} combine single class predictions from different views, alleviating the burden of dealing with heterogeneity and offering greater adaptability to new modalities.

\subsection{Uncertainty Estimation}

Conventional deep learning methods have made notable strides in tackling a diverse set of real-world challenges. However, they continue to fall short in providing insights into the trustworthiness of their predictions. To facilitate dependable decision-making, the incorporation of uncertainty estimation methods is imperative. Currently, some popular methods for uncertainty estimation are as follows:

\subsubsection{Bayesian Neural Networks (BNNs)}

Bayesian Neural Networks (BNNs) use Bayesian inference to estimate the posterior distribution of neural network weights. This approach provides a probability distribution for the forecast, offering insights into the uncertainty associated with the prediction. BNNs also address parametric uncertainty by accounting for variability in model parameters. Bayesian neural networks can better reflect the reliability of model prediction. Key works in this field include Blundell et al. \cite{blundell2015weight}, who introduced a practical Bayesian framework for deep learning by combining dropout regularization with Bayesian modeling, enabling scalable approximation of the posterior distribution. Additionally, BNNs are robust against the issue of overfitting and can be trained effectively on datasets of different scales\cite{kucukelbir2017automatic}. These studies contribute not only theoretically but also offer practical methodologies for implementing BNNs in real-world scenarios, promising enhanced robustness and interpretability in uncertainty estimation of deep neural networks. 

\subsubsection{Monte-Carlo Dropout (MC-Dropout)}
To approximate the posterior inference, the concept of MC-Dropout was introduced by Gal and Ghahramani \cite{gal2016dropout}, who conceptualized dropout as a Bayesian approximation, incorporating uncertainty modeling for neural network weights. Widely applied in computer vision, natural language processing, and beyond, MC-Dropout offers a simple yet effective approach for uncertainty modeling. Zhao\cite{zhao2022pyramid} et al. used MC-Dropout to perform uncertainty estimation on semantic segmentation algorithms.

\subsubsection{Deep Ensemble}
To estimate predictive uncertainty in deep learning models, Lakshminarayanan et al. \cite{lakshminarayanan2017simple} propose the Deep Ensembles (DE) method as a computationally efficient method for quantifying predictive uncertainty in deep neural networks. This approach, unlike Bayesian NNs, is simple to implement, parallelizable, and requires minimal tuning. Furthermore, Molchanov et al. \cite{molchanov2017variational} demonstrated the effectiveness of Deep Ensembles by merging it with MC-Dropout, creating a method that leverages the strengths of both approaches. This integration enhances the accuracy of uncertainty estimates and provides a practical framework for uncertainty-aware decision-making in various applications. Peng et al. \cite{peng2021uncertainty} proposed a Deep Ensemble uncertainty estimation method for the YOLOv3 object detection algorithm, and analyzed the relationship between the uncertainty and confidence of the algorithm.

\subsubsection{Evidential Deep Learning}	 
Sensoy et al.\cite{sensoy2018evidential} introduce a novel approach in deep learning by incorporating evidence theory to more accurately quantify model uncertainty in classification tasks. In comparison to traditional softmax outputs, this method adeptly captures the model's confidence and uncertainty for different classes. Notable works include EvidenceCap \cite{zou2023evidencecap} proposed by Zou et al. is a foundation model for medical image segmentation, that enhances the transparency and reliability of artificial intelligence systems by quantifiably estimating uncertainty. Moreover, Trusted multi-view classification\cite{TMC} proposed by Han et al. introduces a novel paradigm by dynamically integrating different views at an evidence level. By utilizing the Dirichlet distribution and Dempster-Shafer theory, the algorithm jointly leverages multiple views to enhance both classification reliability and robustness. 

\begin{figure*}[t!]
    \centering
    \includegraphics[scale=0.5]{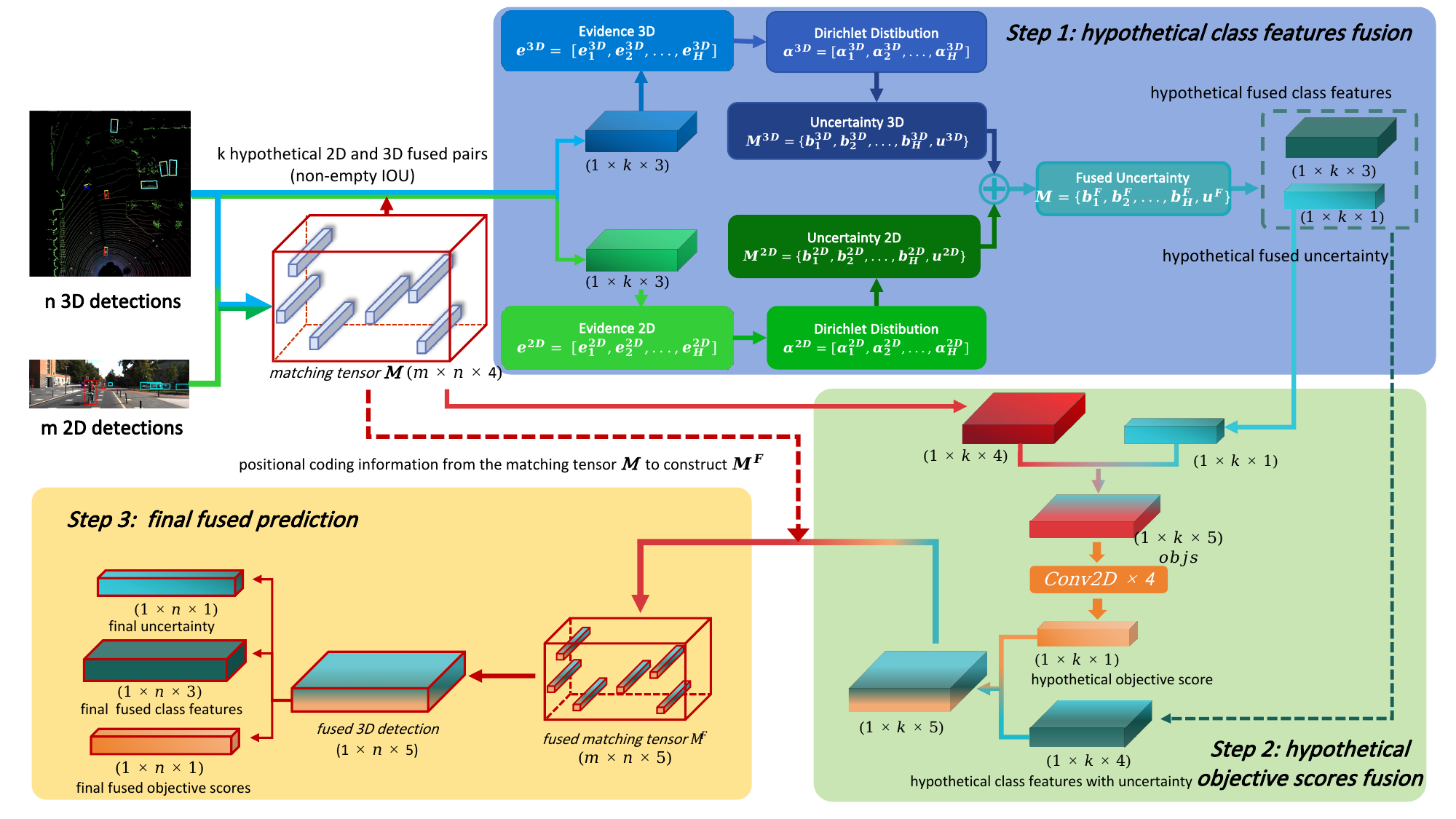}
    \caption{System architecture. In step 1, Each of the $m$ 3D candidates is computed for IOU with each of the $n$ 2D candidates to have $k$ hypothetical fused pairs, and fused class features with uncertainty are obtained based on these pairs. In step 2, the hypothetical objective scores are computed by a 2D CNN and then concatenated with the fused class features with uncertainty which ultimately is used in step 3 to build fused matching tensor $\mathbf{M^F}$ to get the final fused prediction}
    \label{fig1:system_architecture}
\end{figure*}

\section{Methods}

In this section, we introduce our trusted multi-modal fusion object detection system. The overall diagram of our proposed system is shown in Fig. 1.

\subsection{Multi-modal Multi-class Late Fusion (MMLF) architecture}
\subsubsection{Hypothetical Class Features Fusion}
The 2D detector and 3D detector selected for this study are YOLOv8 ~\cite{jocher2023ultralytics} and Complex-YOLO~\cite{cyolo}, respectively, meeting the requirements for multi-class detection suitable for fusion.

The quantities of 3D candidates and 2D candidates are denoted as $m$ and $n$ respectively. $m$ 3D candidates are projected onto the image plane to calculate the IOU with $n$ 2D candidates. This computation leads to the creation of a matching tensor, denoted as $\mathbf{M}$ and having dimensions $(m \times n \times 4)$. Each element within the tensor $\mathbf{M}$  is defined as follows:
\begin{equation}
\mathbf{M_{i,j}} = \left\{ IOU_{i,j}^{\text{3D-2D}}, \mathbf{objs}_{i}^{\text{3D}}, \mathbf{objs}_{j}^{\text{2D}}, \mathbf{dis}_{i}^{\text{3D}} \right\}
\end{equation}

Specifically, $IOU_{i,j}^{\text{3D-2D}}$ represents IOU between the $i_{th}$ projected 3D candidate and the $j_{th}$ 2D candidate, indicating the shared region between them. $\mathbf{objs}_{i}^{\text{3D}}$ and $\mathbf{objs}_{j}^{\text{2D}}$ represent the objective scores of the $i_{th}$ projected 3D candidate and the $j_{th}$ 2D candidate, while $\mathbf{dis}_{i}^{\text{3D}}$ denotes the normalized distance between the $i_{th}$ 3D bounding box and the LiDAR in the xy-plane.

Subsequently, $k$ hypothetical 3D-2D pairs are identified for fusion with non-zero $IOU^{\text{3D-2D}}$. For candidates in these hypothetical pairs, the 3D and 2D class features are regarded as evidence \cite{shafer1992dempster}, and the 3D and 2D evidence $e^{3D} =   \left [ e_{1}^{3D},e_{2}^{3D},...,e_{\text{H}}^{3D}  \right ]$, $e^{2D} =   \left [ e_{1}^{2D},e_{2}^{2D},...,e_{\text{H}}^{2D}  \right ]$, are utilized by subjective logic~\cite{jsang2018subjective} to connect to the parameters of the Dirichlet distribution $\alpha ^{3D} =   \left [  \alpha_{1}^{3D},\alpha_{2}^{3D},...,\alpha_{\text{H}}^{3D}  \right ]$, $\alpha ^{2D} =   \left [  \alpha_{1}^{2D},\alpha_{2}^{2D},...,\alpha_{\text{H}}^{2D}  \right ]$. Consequently, the 3D and 2D belief masses $b ^{3D} =   \left [  b_{1}^{3D},b_{2}^{3D},...,b_{\text{H}}^{3D}  \right ]$, $b ^{2D} =   \left [  b_{1}^{2D},b_{2}^{2D},...,b_{\text{H}}^{2D}  \right ]$, along with the uncertainties $u^{3D}$ and $u^{2D}$, can be calculated using Eq. (2) and Eq. (3), where $H$ is the number of classes (e.g., $H=3$ in the KITTI Dataset), and Dirichlet strength is denoted by $S=\sum_{i}^{H} \alpha_i$.

\begin{equation}
b_h = \frac{e_h}{S} = \frac{\alpha _h - 1}{S} \text{(h ranges from 1 to H) }
\end{equation}

\begin{equation}
    u = \frac{H}{S}
\end{equation}
To obtain the fused mass $M ^{F} =   \left \{ b_{1}^{F},b_{2}^{F},...,b_{\text{H}}^{F},u^F  \right \}$, Dempster’s combination rule ~\cite{sentz2002combination}~\cite{josang2012interpretation} is employed to calculate the fused belief mass $\left [  b_{1}^{F},b_{2}^{F},...,b_{\text{H}}^{F} \right ]$  and fused uncertainty $u^F$ in Eq. (4) and Eq. (5), where $C =  {\textstyle \sum_{i\ne{j}}^{}} b_{i}^{3D} b_{j}^{2D} $  measures the conflict between 2D and 3D class detection.

\begin{equation}
b_{h}^{F}=\frac{1}{1-C}\left (b_{h}^{3D}b_{h}^{2D}+b_{h}^{3D}u^{2D}+b_{h}^{2D}u^{3D}   \right ) 
\end{equation}
\begin{equation}
u^{F}=\frac{1}{1-C}u^{3D}u^{2D}
\end{equation}

\subsubsection{Hypothetical Objective Scores Fusion}

For each of the \(k\) hypothetical 3D-2D pairs, the fused uncertainty $u^F$, which represents the degree of class conflict between 2D and 3D modalities, is incorporated based on the matching tensor $\mathbf{M}$ to construct a \(1 \times k \times 5\) objective score tensor $\mathbf{objs}$, as shown in Eq. (6):

\begin{equation}
\mathbf{objs} = \left\{ IOU_{i,j}^{\text{3D-2D}}, \mathbf{objs}_{i}^{\text{3D}}, \mathbf{objs}_{j}^{\text{2D}}, \mathbf{dis}_{i}^{\text{3D}}, \mathbf{u}_{i,j}^{F} \right\}
\end{equation}

To fuse the $\mathbf{objs}$ to get a hypothetical objective score, we use a set of 2D convolution layers with a kernel size of (1,1) and a stride of 1 to convolve the third channel of $\mathbf{objs}$ from $1 \times k \times 5$ to $1 \times k \times 18$, from $1 \times k \times 18$ to $1 \times k \times 36$, and finally from $1 \times k \times 36$ to $1 \times k \times 1$ to obtain the hypothetical objective score vector with a size of $1 \times k \times 1$.
After that, the hypothetical objective score vector ($1 \times k \times 1$) is concatenated with the hypothetical class features along with uncertainty ($1 \times k \times 4$).

\subsubsection{Final Fused Prediction}
With the concatenated vector ($1 \times k \times 5$) based on the positional coding from the matching tensor $\mathbf{M}$, we proceed to construct the fused matching tensor $\mathbf{M^F}$. The tensor with the highest objective score in the second dimension is then selected using the Max function to obtain the final fused 3D detection, encompassing the final fused objective scores, final fused class features, and final uncertainty. Following the application of Non-Maximum Suppression (NMS) and filtering out detection results with high uncertainty, a more confident final prediction is obtained.
\subsection{Non-matching Scenarios}
For any projected 3D candidate that has no 2D candidate intersected, we still retain the last 3D-2D pair for this 3D candidate, assigning a 2D confidence score and to be -10. Furthermore, the fused belief mass $   \left [  b_{1}^{F},b_{2}^{F},...,b_{\text{H}}^{F}  \right ]  $ and fused uncertainty $u^F$ are replaced by the original 3D modality belief mass $  \left [  b_{1}^{3D},b_{2}^{3D},...,b_{\text{H}}^{3D}  \right ]  $ and 3D uncertainty $u^{3D}$. These two adjustments are made to ignore illogical fusion and preserve information from 3D views. In addition, we need to calculate separate loss functions for the matched and fused results, as well as the unmatched results.
\subsection{Uncertainty Estimation}
Initially, we performed hypothetically fusion based on IOU information from matching tensor $\mathbf{M}$ to get the hypothetical fused uncertainty, adding the information of conflicts between two different modalities to objective score tensor $\mathbf{objs}$ in the hypothetical objective scores fusion to reduce illogical fusion to be selected after convolution. Subsequently, in the final fused predictions of the network, by combining with final fused objective scores, we enhanced the representational information and interpretability of predictions, which allows us to distinguish the network's confidence in both object localization and object categorization. Through subsequent filtering out detections with high uncertainty scores, unreasonable results are removed to achieve more reliable decisions.
\subsection{Loss Function}

For the final fused 3D detection, binary cross-entropy (BCE) loss $L_{BCE}$ is applied for the objective loss $Loss_{Obj}$ with objective-mask and no-objective-mask as described:
\begin{equation}
 Loss_{Obj} =  L_{\textit{BCE}}(\text{\textit{obj\_mask}}) + L_{\textit{BCE}}(\text{\textit{no\_obj\_mask}})
\end{equation}

Additionally, for the final fused 3D detection, we use the sample-specific loss  $ L_{ssl} $ from \cite{TMC} for fused object classification and adjust it slightly for non-matching scenarios. The overall loss is expressed as follows:
\begin{equation}
\begin{aligned}
& \textstyle Loss = \sum_{i=1}^{N}\left[ L_{ssl}(\alpha_i^F) + L_{ssl}(\alpha_i^{3D}) + L_{ssl}(\alpha_i^{2D}) \right] \\
& \quad + \sum_{i=1}^{N} L_{ssl}(\alpha_i^{3D})_{\text{\textit{no-matching}}} + Loss_{Obj}
\end{aligned}
\end{equation}

To optimize the network during training, we utilize the Adam optimizer with a learning rate of 0.003 for the hypothetical class features fusion and employ the Stochastic Gradient Descent (SGD) optimizer with the same learning rate for the hypothetical objective scores fusion
\section{Experiment and Results}
\subsection{Experiments Setup}
We applied the MMLF method to evaluate object detection on the KITTI dataset \cite{geiger2012we} for 2D object detection, 2D Orientation, 3D object detection and bird’s eye view (BEV), encompassing both the split validation dataset and the official test dataset. In detail, we split the original 7481 training samples into 6000 training samples and 1481 validation samples, and KITTI official toolkit is used for validation. For 7518 testing samples, we submit the inference result to the official KITTI server. The data values are represented as Average Precision (AP), with the unit being percentage (\%).
For 3D object detection, the detectors utilized are Complex-YOLO v3 and v4, which exclusively process LiDAR point cloud data. Simultaneously, the 2D object detection was performed using YOLO v8, focusing on objects visible in the image plane.

\subsection{KITTI Validation Dataset Results}
The initial 3D results from the detectors Complex-YOLO v3 and v4 are labeled as Ori-v3 and Ori-v4, serving as the baseline. The fused results are represented as Fuse-v3 and Fuse-v4, corresponding to Complex-YOLO v3 fused with YOLO v8 and Complex-YOLO v4 fused with YOLO v8. 
Furthermore, the improvement with yellow highlighted is expressed as the difference between the fused result and the original 3D results. At the NMS stage, the objective confidence threshold is 0.95, and the NMS threshold is 0.4. 

In Table I, we present the evaluation results on the validation dataset. For 2D bounding box detection, significant improvements are observed across all difficulty levels (Easy, Moderate, Hard) for car, pedestrian, and cyclist categories when comparing Fuse-v3 and Fuse-v4 with their respective original 3D results. The improvements range from +4.68\% to +19.57\%, showcasing the effectiveness of the fusion approach.
In terms of 2D bounding box orientation, similar positive trends are observed, with improvements ranging from +5.59\% to +19.41\% across different categories and difficulty levels.

\begin{table}[h]
\caption{KITTI Validation Dataset Performance Comparison}
\small
\centering
\resizebox{\columnwidth}{!}{%
\begin{tabular}{l|ccc|ccc|ccc}
\textbf{Methods} & \multicolumn{3}{c|}{\textbf{car}} & \multicolumn{3}{c|}{\textbf{Pedestrian}} & \multicolumn{3}{c}{\textbf{Cyclist}} \\ \cline{2-10}
 & \textbf{Easy} & \textbf{Mod.} & \textbf{Hard} & \textbf{Easy} & \textbf{Mod.} & \textbf{Hard} & \textbf{Easy} & \textbf{Mod.} & \textbf{Hard} \\ \hline
\multicolumn{10}{c}{\textbf{eval 2D Detection AP (\%)}} \\ \hline
Ori-v3 & \customDataRow{56.85} {50.94} {53.23}{24.51} {26.57} {24.50}{54.38} {43.52} {44.14} \\
fuse-v3 & \customDataRow{\textbf{65.31}} {\textbf{57.48}} {\textbf{58.89}}{\textbf{38.43}} {\textbf{37.83}} {\textbf{37.77}}{\textbf{66.40}} {\textbf{54.54}} {\textbf{54.35}}\\\hline
\rowcolor{yellow}
Improvement & \customDataRow{\textbf{+8.46}}{\textbf{+6.54}}{\textbf{+5.66}}{\textbf{+13.92}}{\textbf{+11.26}}{\textbf{+13.27}}{\textbf{+12.02}}{\textbf{+11.02}}{\textbf{+10.21}}\\\hline
Ori-v4 & \customDataRow{56.46} {51.72} {54.74}{49.17} {52.11} {53.50}{55.19} {43.99} {44.25} \\
fuse-v4 & \customDataRow {\textbf{67.80}} {\textbf{59.12}} {\textbf{60.48}}{\textbf{68.74}} {\textbf{62.95}} {\textbf{62.84}} {\textbf{66.92}} {\textbf{61.32}} {\textbf{55.36}}\\ \hline
\rowcolor{yellow} 
Improvement & \customDataRow{\textbf{+11.34}} {\textbf{+7.40}} {\textbf{+5.74}} {\textbf{+19.57}} {\textbf{+10.64}} {\textbf{+8.46}} {\textbf{+12.40}} {\textbf{+7.59}} {\textbf{+4.94}}\\\hline
\multicolumn{10}{c}{\textbf{eval 2D Orientation AP (\%)}} \\ \hline
Ori-v3 & \customDataRow{56.84} {50.87} {53.16}{24.09} {26.12} {24.19}{53.66} {43.30} {43.93} \\
fuse-v3 & \customDataRow{\textbf{65.30}} {\textbf{57.46}} {\textbf{58.84}}{\textbf{35.31}} {\textbf{34.75}} {\textbf{34.94}}{\textbf{66.26}} {\textbf{54.43}} {\textbf{54.24}}\\\hline
\rowcolor{yellow}
Improvement & \customDataRow{\textbf{+8.46}}{\textbf{+6.59}}{\textbf{+5.68}}{\textbf{+11.22}}{\textbf{+8.63}}{\textbf{+10.75}}{\textbf{+12.60}}{\textbf{+11.13}}{\textbf{+10.31}}\\\hline
Ori-v4 & \customDataRow{56.45} {51.68} {54.70}{47.71} {50.57} {51.83}{55.10} {43.94} {44.19} \\
fuse-v4 & \customDataRow{\textbf{67.79}} {\textbf{59.09}} {\textbf{60.42}}{\textbf{67.12}} {\textbf{61.26}} {\textbf{60.35}}{\textbf{66.84}} {\textbf{61.25}} {\textbf{55.30}}
 \\ \hline
\rowcolor{yellow}
Improvement & \customDataRow{\textbf{+11.34}} {\textbf{+7.41}} {\textbf{+5.72}} {\textbf{+19.41}} {\textbf{+10.69}} {\textbf{+8.52}} {\textbf{+11.74}} {\textbf{+17.31}} {\textbf{+11.11}}\\\hline
\multicolumn{10}{c}{\textbf{BEV AP (\%)}} \\ \hline
Ori-v3 & \customDataRow{89.68} {89.02} {89.12}{51.96} {54.25} {55.84}{63.94} {71.47} {71.95}\\
fuse-v3 & \customDataRow  {\textbf{89.92}} {\textbf{89.30}} {\textbf{89.44}} {\textbf{64.67}} {\textbf{65.35}} {\textbf{66.34}} {\textbf{69.29}} {\textbf{77.29}} {\textbf{77.22}}\\\hline
\rowcolor{yellow}
Improvement & \customDataRow{\textbf{+0.24}}{\textbf{+0.28}}{\textbf{+0.32}}{\textbf{+12.71}}{\textbf{+11.10}}{\textbf{+10.50}}{\textbf{+5.35}}{\textbf{+5.82}}{\textbf{+5.27}}\\\hline
Ori-v4 & \customDataRow{89.46} {89.33} {89.43}{56.71} {54.57} {55.88} {75.25} {76.51} {76.41}\\
fuse-v4 & \customDataRow{\textbf{90.06}} {\textbf{89.61}} {\textbf{89.61}}{\textbf{73.75}} {\textbf{69.26}} {\textbf{69.08}}{\textbf{76.72}} {\textbf{77.25}} {\textbf{77.40}} \\ \hline
\rowcolor{yellow}
Improvement & \customDataRow{\textbf{+0.60}} {\textbf{+0.28}} {\textbf{+0.18}} {\textbf{+17.04}} {\textbf{+14.69}} {\textbf{+13.20}} {\textbf{+1.47}} {\textbf{+0.74}} {\textbf{+1.99}}
 \\\hline
\multicolumn{10}{c}{\textbf{eval 3D detection AP (\%)}} \\ \hline
Ori-v3 & \customDataRow{29.50} {27.74} {29.70} {30.34} {28.68} {30.61} {19.95} {19.87} {20.82}\\
fuse-v3 & \customDataRow{\textbf{38.43}} {\textbf{34.62}} {\textbf{35.91}} {\textbf{64.67}} {\textbf{65.35}} {\textbf{66.34}}{\textbf{69.29}} {\textbf{77.29}} {\textbf{77.22}}\\\hline
\rowcolor{yellow}
Improvement & \customDataRow{\textbf{+8.93}}{\textbf{+6.88}}{\textbf{+6.21}}{\textbf{+34.33}}{\textbf{+36.67}}{\textbf{+35.73}}{\textbf{+49.34}}{\textbf{+57.42}}{\textbf{+56.40}}\\hline
Ori-v4 & \customDataRow{37.04} {36.13} {39.23}{43.06} {40.77} {42.10} {39.16} {29.87} {30.47}\\
fuse-v4 & \customDataRow{\textbf{50.75}} {\textbf{44.62}} {\textbf{45.50}} {\textbf{58.38}} {\textbf{54.07}} {\textbf{49.25}}{\textbf{51.15}} {\textbf{40.91}} {\textbf{40.75}}\\\hline
\rowcolor{yellow}
Improvement & \customDataRow{\textbf{+13.71}} {\textbf{+8.49}} {\textbf{+6.27}} {\textbf{+15.32}} {\textbf{+13.30}} {\textbf{+7.15}} {\textbf{+11.99}} {\textbf{+11.04}} {\textbf{+10.28}}
 \\\hline
\end{tabular}
}

\end{table}

For BEV bounding boxes, improvements are evident in the Easy and Moderate scenarios, ranging from +0.18\% to +17.04\%. However, in car detection, improvements are less pronounced.

In the evaluation of 3D bounding boxes, Fuse-v3 and Fuse-v4 consistently outperform Ori-v3 and Ori-v4, with improvements ranging from +6.27\% to +16.72\%. These enhancements underscore the effectiveness of the fusion method in refining 3D detection accuracy.

Overall, the fusion approach consistently demonstrates superiority over the baseline in various evaluation metrics, with improvements ranging from +0.18\% to +19.57\%. These results highlight the efficacy of the fusion technique in enhancing object detection across different scenarios and categories on the validation dataset.

\begin{table}[H]
\caption{KITTI Test Dataset Performance Comparison}
\small
\centering
\resizebox{\columnwidth}{!}{%
\begin{tabular}{l|ccc|ccc|ccc}
\textbf{Methods} & \multicolumn{3}{c|}{\textbf{car}} & \multicolumn{3}{c|}{\textbf{Pedestrian}} & \multicolumn{3}{c}{\textbf{Cyclist}} \\ \cline{2-10}
 & \textbf{Easy} & \textbf{Mod.} & \textbf{Hard} & \textbf{Easy} & \textbf{Mod.} & \textbf{Hard} & \textbf{Easy} & \textbf{Mod.} & \textbf{Hard} \\ \hline
\multicolumn{10}{c}{\textbf{Test 2D Detection AP (\%)}} \\ \hline
Ori-v3 & \customDataRow{47.02} {37.02} {37.01} {10.06} {7.07} {7.19}  {28.73} {21.08} {18.32}  \\
Fuse-v3 & \customDataRow{\textbf{58.07}} {\textbf{44.00}} {\textbf{43.20}} {\textbf{21.69}} {\textbf{14.47}} {\textbf{13.03}}
{\textbf{36.64}} {\textbf{26.94}} {\textbf{24.90}}

  \\ \hline
\rowcolor{yellow} 
Improvement & \customDataRow{\textbf{+11.05}}{\textbf{+6.98}}{\textbf{+6.19}}{\textbf{+11.63}}{\textbf{+7.40}}{\textbf{+5.84}}{\textbf{+7.91}}{\textbf{+5.86}}{\textbf{+6.58}}
 \\\hline
Ori-v4 & \customDataRow{44.72} {38.36} {37.88}{27.17} {19.97} {18.74}{36.21} {26.69} {23.90}  \\
Fuse-v4 & \customDataRow{\textbf{61.58}} {\textbf{45.67}} {\textbf{44.47}}{\textbf{39.56}} {\textbf{28.77}} {\textbf{26.77}}
{\textbf{21.32}} {\textbf{15.01}} {\textbf{13.50}}

  \\ \hline
\rowcolor{yellow} 
Improvement & \customDataRow{\textbf{+16.86}}{\textbf{+7.31}}{\textbf{+6.59}}{\textbf{+12.39}}{\textbf{+8.80}}{\textbf{+8.03}}{\textbf{+15.11}}{\textbf{+11.32}}{\textbf{+9.60}} \\\hline

\multicolumn{10}{c}{\textbf{Test 2D Orientation AP (\%)}} \\ \hline
Ori-v3 & \customDataRow{47.00} {36.90} {36.85}  {6.27} {4.42} {4.48} {27.80} {19.42} {16.95}  \\
Fuse-v3 & \customDataRow {\textbf{58.03}} {\textbf{43.81}} {\textbf{42.93}}{\textbf{13.48}} {\textbf{9.13}} {\textbf{8.13}}
{\textbf{34.35}} {\textbf{23.81}} {\textbf{22.02}}

\\ \hline
\rowcolor{yellow}
Improvement & \customDataRow{\textbf{+11.03}}{\textbf{+6.91}}{\textbf{+6.08}}{\textbf{+7.21}}{\textbf{+4.71}}{\textbf{+3.65}}{\textbf{+6.55}}{\textbf{+4.39}}{\textbf{+5.07}}
 \\\hline

Ori-v4 & \customDataRow{44.60} {38.12} {37.60}{21.33} {15.37} {14.27}{34.59} {25.47} {22.80}\\
Fuse-v4 & \customDataRow{\textbf{61.32}} {\textbf{45.41}} {\textbf{44.17}} {\textbf{30.96}} {\textbf{22.18}} {\textbf{20.38}}
{\textbf{38.03}} {\textbf{28.53}} {\textbf{25.12}}

\\ \hline
\rowcolor{yellow}
Improvement & \customDataRow{\textbf{+16.72}}{\textbf{+7.29}}{\textbf{+6.57}}{\textbf{+9.63}}{\textbf{+6.81}}{\textbf{+6.11}}{\textbf{+3.44}}{\textbf{+3.06}}{\textbf{+2.32}} \\\hline

\multicolumn{10}{c}{\textbf{Test BEV AP (\%)}} \\ \hline
Ori-v3 & \customDataRow{\textbf {83.02}} {\textbf{74.39}} {66.34}{16.13} {12.62} {11.55} {27.53} {19.22} {17.65}\\ 
Fuse-v3 & \customDataRow{80.24} {72.91} {\textbf{67.26}}{\textbf{20.16}} {\textbf{15.17}} {\textbf{13.87}}
 {\textbf{32.24}} {\textbf{26.42}} {\textbf{23.34}}

 \\ \hline
\rowcolor{yellow}
Improvement & \customDataRow{-2.78} {-1.48}{\textbf{+0.92}}{\textbf{+4.03}}{\textbf{+2.55}}{\textbf{+2.32}}{\textbf{+4.71}}{\textbf{+7.20}}{\textbf{+5.69}}
  \\\hline
Ori-v4 & \customDataRow{81.88} {75.38} {69.42}{25.00} {18.42} {17.20}{39.47} {30.80} {27.62}\\ 
Fuse-v4 & \customDataRow{\textbf{84.23}} {\textbf{77.99}} {\textbf{71.74}} {\textbf{33.46}} {\textbf{25.96}} {\textbf{23.15}}
{\textbf{41.28}} {\textbf{34.98}} {\textbf{29.68}}
 \\ \hline
\rowcolor{yellow}
Improvement & \customDataRow{\textbf{+2.35}}{\textbf{+2.61}}{\textbf{+2.32}}{\textbf{+8.46}}{\textbf{+7.54}}{\textbf{+5.95}}{\textbf{+1.81}}{\textbf{+4.18}}{\textbf{+2.06}} \\\hline

\multicolumn{10}{c}{\textbf{Test 3D bounding boxes AP (\%)}} \\ \hline
Ori-v3 & \customDataRow{16.25} {13.39} {13.42}{7.22} {5.33} {4.64} {8.57} {5.43} {5.46}  \\
Fuse-v3 & \customDataRow{\textbf{21.51}} {\textbf{16.23}} {\textbf{15.92}}{\textbf{12.08}} {\textbf{8.57}} {\textbf{7.56}}
{\textbf{11.95}} {\textbf{8.49}} {\textbf{7.67}}

\\\hline
\rowcolor{yellow}
Improvement & \customDataRow{\textbf{+5.26}}{\textbf{+2.84}}{\textbf{+2.50}}{\textbf{+4.86}}{\textbf{+3.24}}{\textbf{+2.92}}{\textbf{+3.38}}{\textbf{+3.06}}{\textbf{+2.21}}
  \\\hline

Ori-v4 & \customDataRow{20.41} {18.26} {18.08}{13.77} {10.23} {9.24}{13.85} {9.10} {8.53}  \\
Fuse-v4 & \customDataRow{\textbf{31.32}} {\textbf{23.82}} {\textbf{22.94}} {\textbf{21.32}} {\textbf{15.01}} {\textbf{13.50}}
{\textbf{17.90}} {\textbf{12.16}} {\textbf{10.55}}

\\\hline
\rowcolor{yellow}
Improvement & \customDataRow{\textbf{+10.91}}{\textbf{+5.56}}{\textbf{+4.86}}{\textbf{+7.55}}{\textbf{+4.78}}{\textbf{+4.26}}{\textbf{+4.05}}{\textbf{+3.06}}{\textbf{+2.02}}\\\hline

\end{tabular}
}

\end{table}

\subsection{KITTI Test Dataset Results}
In Table II, the test dataset evaluation results are presented. For 2D bounding box detection, substantial improvements are observed across all difficulty levels (Easy, Moderate, Hard) for car, pedestrian, and cyclist categories when comparing Fuse-v3 and Fuse-v4 with their respective original 3D results. The improvements range from +5.84\% to +16.86\%, highlighting the effectiveness of the fusion approach.
In terms of 2D bounding box orientation, similar positive trends are observed, with improvements ranging from +2.32\% to +16.72\% across different categories and difficulty levels.

For BEV bounding boxes, improvements are evident in the majority of cases. Moderate, and Hard scenarios, ranging from +0.92\% to +8.46\%. However, in car detection of oriv3,  there is a slight loss of performance. 

In the evaluation of 3D bounding boxes, Fuse-v3 and Fuse-v4 consistently outperform Ori-v3 and Ori-v4, with improvements ranging from +2.02\% to +10.91\%. These enhancements underscore the effectiveness of the fusion method in refining 3D detection accuracy.

In conclusion, the fusion approach demonstrates superiority over the baseline in various evaluation metrics on the test dataset, with improvements ranging from +0.92\% to +16.72\%. These results highlight the efficacy of the fusion technique in enhancing object detection across different scenarios and categories.
\subsection{Uncertainty Analysis}
\subsubsection{Quantitive Results}
The detailed calculation of uncertainty scores is shown in Eq. (3) and Eq. (5). To more intuitively understand the uncertainty score, an example detection of a car with original 3d evidence [22.29, 0.01, 0.01] will get the uncertainty score of 0.1186, and the fused evidence [136.38, 0.01, 0.01] will get the uncertainty score of 0.0215.

\begin{table}[h]
\caption{Average uncertainty score comparison of v3}
\centering
\begin{tabular}{|c|c|c|c|}
\hline
\diagbox{Method}{Class} & car & pedestrian & cyclist \\ \hline
Ori-v3& 0.11827 & 0.24373 & 0.23636 \\ \hline
Fuse v3& \textbf{0.02692} & \textbf{0.05354} & \textbf{0.07354} \\ \hline
\end{tabular}
\label{tab:my_table}
\end{table}

\begin{table}[h]
\caption{Average uncertainty score comparison of v4}
\centering
\begin{tabular}{|c|c|c|c|}
\hline
\diagbox{Method}{Class} & car & pedestrian & cyclist \\ \hline
Ori-v4 & 0.11881 & 0.25032 & 0.23954 \\ \hline
Fuse-v4 & \textbf{0.02737} & \textbf{0.05797} & \textbf{0.07926} \\ \hline
\end{tabular}
\label{tab:my_table}
\end{table}
For Table III (v3 comparison):
The average uncertainty score for cars in the original method (Ori-v3) is 0.11827, which decreases significantly to 0.02692 in the fused method (Fuse v3). This indicates that the fusion approach reduces uncertainty in car detection. Similarly, for pedestrians, Ori-v3 has an average uncertainty score of 0.24373, which drops to 0.05354 with the fused method. Again, a notable reduction in uncertainty is observed. For cyclists, the uncertainty score also decreases from 0.23636 (Ori-v3) to 0.07354 (Fuse v3), indicating improved confidence in cyclist detection after fusion.

For Table IV (v4 comparison):
The average uncertainty scores for cars, pedestrians, and cyclists in the original (Ori-v4) method are 0.11881, 0.25032, and 0.23954 respectively, decreasing to 0.02737, 0.05797, and 0.07926 in the fused (Fuse-v4) method. This demonstrates a consistent pattern of reduced uncertainty in object detection tasks after fusion, enhancing confidence and reliability.

Overall, both versions (v3 and v4) show a consistent pattern where the fusion approach leads to a substantial decrease in uncertainty across all classes (car, pedestrian, and cyclist). This suggests that the fusion methodology is effective in enhancing the model's confidence and reliability in object detection tasks.

\subsubsection{Qualitative results}
\begin{figure}[h]
    \centering
    \setlength{\abovecaptionskip}{0.1cm}
    \includegraphics[width=\columnwidth,scale=0.6]{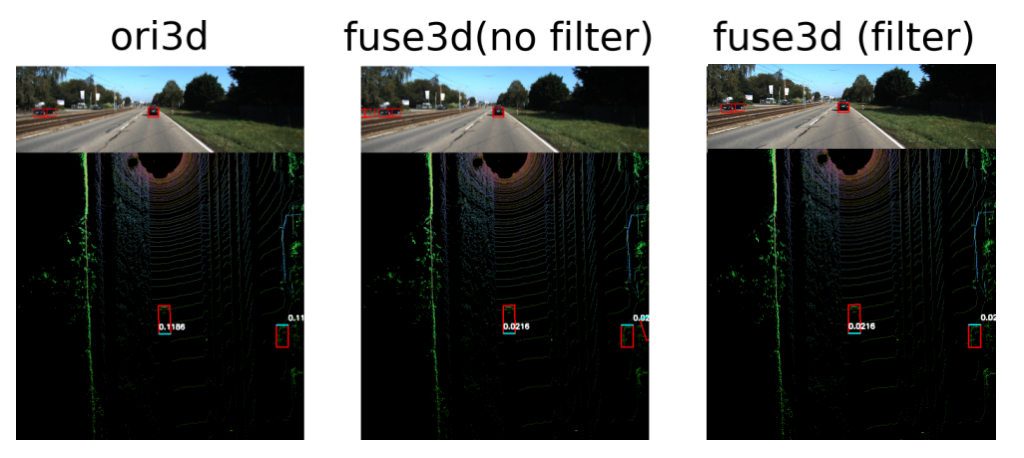}
    \caption{Example of uncertainty filtering. }
    \label{fig2:uex1}
    \setlength{\belowcaptionskip}{0cm}
\end{figure}

 As depicted in Fig. \ref{fig2:uex1}, the fused 3D model mistakenly identified a car on the left side of the road with a high fused uncertainty score of 0.1201, so we filtered it. This further improves the reliability of our detection model.
\subsection{Summary}
\begin{figure}[h]
    \centering
    \centering
        \includegraphics[angle=-90, width=\columnwidth]{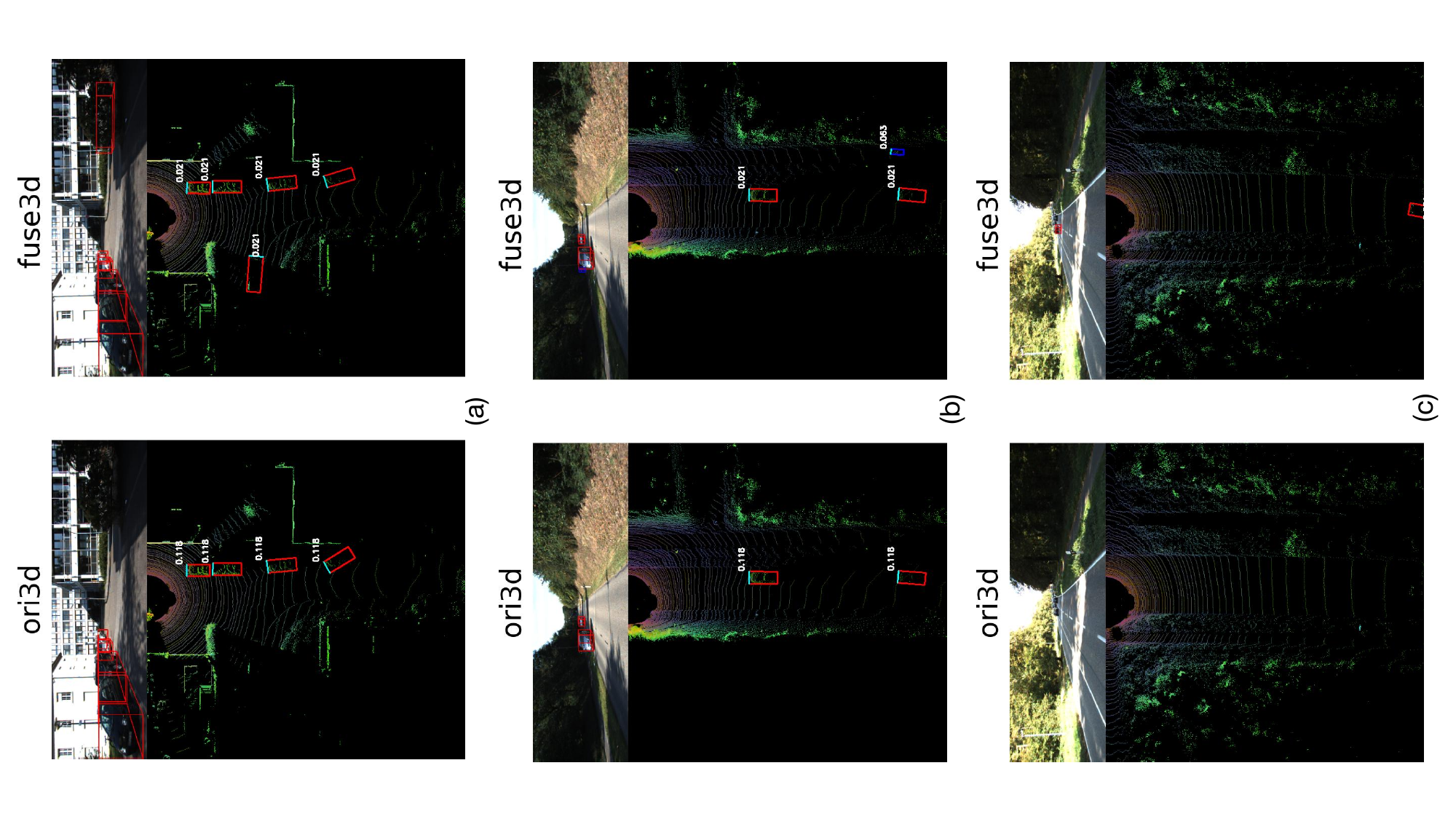}
        \setlength{\abovecaptionskip}{0.5cm}
        \caption{Visualized results of our MMLF on KITTI validation set. In the 3D BEV representation, the numbers adjacent to the bounding boxes indicate uncertainty. Our model has significantly reduced the original detection uncertainty and demonstrated an improvement in detection capability. (a) Demonstrates the enhanced detection capability of our fused model in the presence of occluded objects.
        (b) Illustrates the improved detection ability of our model for small objects.
        (c) Indicates that our model partially addresses the issue of weak long-range object detection inherent in 3D detectors.}
        \label{fig:sub2}
        
        \setlength{\belowcaptionskip}{0cm}
\end{figure}

The model's significance in autonomous driving detection is multifaceted. Firstly, it significantly enhances detection and positioning accuracy by seamlessly fusing 2D and 3D information, thereby improving the overall performance of the sensing module. Notably, it excels in accurately predicting pedestrian directions, a critical capability for autonomous driving systems to anticipate pedestrian behavior and enhance safety in diverse traffic scenarios. Secondly, the model demonstrates strong generalization ability, showcasing robust performance not only on verification sets but also on test sets. This underscores its adaptability across various scenarios and conditions, providing a reliable solution for real-world deployment in autonomous driving systems. Moreover, the model seamlessly integrates information from a wider range of detection patterns, efficiently fusing object-level information from multiple modalities without altering its original structure. This adaptability enables the model to capture richer environmental representations in autonomous driving scenarios. Additionally, the model incorporates uncertainty estimation, enabling the evaluation of detection reliability based on uncertainty scores—a critical aspect for ensuring safe and efficient driving systems. Furthermore, visualized results in Fig. \ref{fig:sub2} provide an intuitive demonstration of the model's effectiveness in real-world scenarios, highlighting its practical applicability in autonomous driving contexts.
% \subsection{Qualitative results}

% \begin{figure}[H]
% \centering
% \includegraphics[scale=0.2]{compare_image006537.png}
% \caption{Title of picture}
% \end{figure*}

\section{Conclusion and Future Work}

In this paper, we introduce an MMLF model designed to effectively integrate object-level information and uncertainty analysis from diverse modalities. Experimental investigations involve the fusion of Complex-YOLOv3/v4, and YOLOv8, demonstrating substantial performance enhancements on the KITTI validation dataset and official test dataset. We have also explored uncertainty estimation, reducing the uncertainty scores of the original model and further enhancing the reliability of our detection results by filtering out detections with high uncertainty. The proposed approach augments detection capabilities across various modalities while preserving the structural integrity of the original detectors. The model exhibits scalability, providing a foundation for future exploration into fusing additional modalities. Furthermore, as part of our prospective work, we aim to address the fusion handling of objects with non-zero IOU overlaps.

\section*{Acknowledgement}
This work was supported in part by the National Key Research and Development Program of China (No. 2022YFB2503004), the Fundamental Research Funds for the Central Universities (NO. ZYGX2022J017), and the Opening Project of International Joint Research Center for Robotics and Intelligence System of Sichuan Province (Grant JQZN2023-005).

\bibliographystyle{IEEEtran}
\small\bibliography{reference}

\end{document}